\documentclass[lettersize,journal]{IEEEtran}
\usepackage{amsmath,amsfonts}
\usepackage{algorithmic}
\usepackage{algorithm}
\usepackage{array}
\usepackage[caption=false,font=normalsize,labelfont=sf,textfont=sf]{subfig}
\usepackage{textcomp}
\usepackage{stfloats}
\usepackage{url}
\usepackage{verbatim}
\usepackage{graphicx}
\usepackage{cite}
\usepackage{amsmath}
\usepackage{amssymb}
\usepackage{booktabs}
\usepackage{color}
\usepackage{multirow}
\usepackage{enumerate}
\newcommand{\vb}[1]{\mathbf{#1}}
\newcommand{\homec}[1]{\overline{\mathbf{#1}}}
\newcommand{\dlt}[1]{}

\begin{document}

\title{Exploiting Correspondences with All-pairs Correlations for Multi-view Depth Estimation}

\author{Kai Cheng,
Hao Chen, Wei Yin,
Guangkai Xu,
Xuejin Chen
\thanks{K. Cheng, G. Xu, X. Chen are with the National Engineering Laboratory for Brain inspired Intelligence Technology and Application, University of Science and Technology of China, Hefei 230026, China. (e-mail: chengkai21@mail.ustc.edu.cn, xugk@mail.ustc.edu.cn, xjchen99@ustc.edu.cn)}
\thanks{H. Chen is with the Computer Graphics and Computer Assisted Design Laboratory, Zhejiang University, Hangzhou 310058, China. (e-mail: haochen.cad@zju.edu.cn)}
\thanks{W. Yin is with DJI Technology. (e-mail: yvanwy@outlook.com)}
\thanks{Part of this work was done when KC, GX, HC were with Huawei. WY was with the University of Adelaide, Australia.}}



\maketitle

\begin{abstract}
 Multi-view depth estimation plays a critical role in reconstructing and understanding the 3D world. Recent learning-based methods have made significant progress in it. 
 However, multi-view depth estimation is fundamentally a correspondence-based optimization problem, but previous learning-based methods mainly rely on predefined depth hypotheses to build correspondence as the cost volume and implicitly regularize it to fit depth prediction, deviating from the essence of iterative optimization based on stereo correspondence. Thus, they suffer unsatisfactory precision and generalization capability. 
In this paper, we are the first to explore more general image correlations to establish correspondences dynamically for depth estimation. We design a novel iterative multi-view depth estimation framework mimicking the optimization process, which consists of 1) a correlation volume construction module that models the pixel similarity between a reference image and source images as all-to-all correlations; 2) a flow-based depth initialization module that estimates the depth from the 2D optical flow; 3) a novel correlation-guided depth refinement module that reprojects points in different views to effectively fetch relevant correlations for further fusion and integrate the fused correlation for iterative depth update. Without predefined depth hypotheses, the fused correlations establish multi-view correspondence in an efficient way and guide the depth refinement heuristically. We conduct sufficient experiments on ScanNet, DeMoN, ETH3D, and 7Scenes to demonstrate the superiority of our method on multi-view depth estimation and its best generalization ability. 
\end{abstract}


\section{Introduction}
\label{sec:intro}
\begin{figure}
  \centering
  \includegraphics[width=\linewidth]{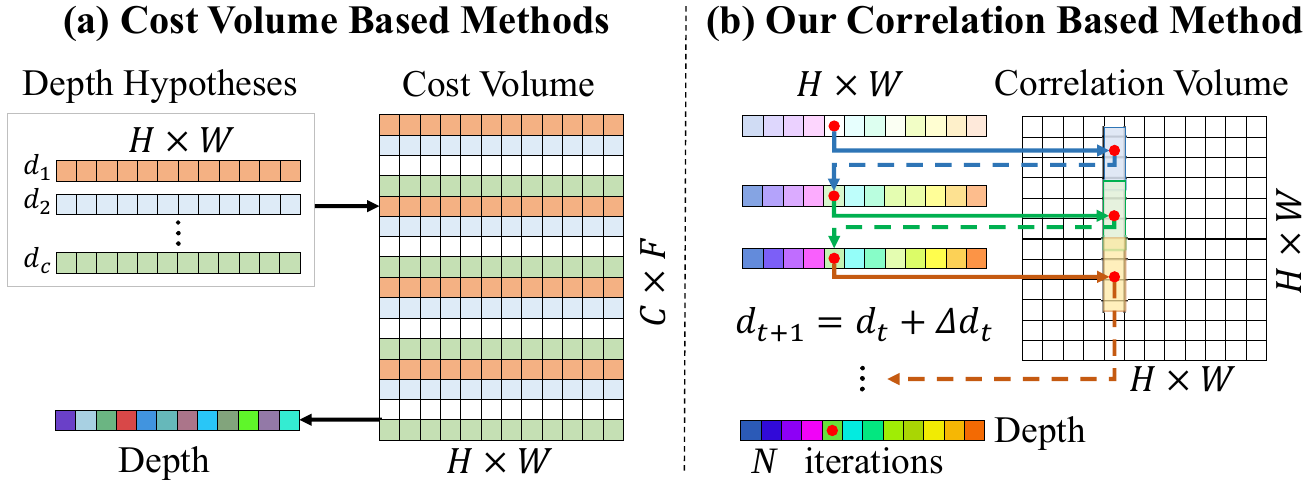}
  \caption{Comparison of our correlation-based method with cost-volume-based methods. Without using predefined depth hypotheses, we iteratively update the depth for each point based on the correlations which express the confidence of the correspondence from the estimated depth.}
  \label{fig:intro}
\end{figure}

\IEEEPARstart{M}{ulti-view} depth estimation is a fundamental problem in 3D vision. It aims to recover the pixel depths with known camera parameters and provides geometric information for many vision tasks, such as 3D reconstruction \cite{schonberger2016pixelwise}, 3D detection \cite{song2014sliding}, and semantic segmentation \cite{couprie2013indoor}. 
%
Reviewing previous methods of multi-view depth estimation, it is crucial to establish point correspondences among different views. Traditional methods \cite{campbell2008using, collins1996space, hosni2012fast, hirschmuller2007stereo} use handcrafted similarity metrics, e.g., the sum of absolute differences and normalized cross-correlation, to describe the point correspondence between image pairs. However, those handcrafted local features are sensitive to illumination variation, low texture, and repetitive patterns.
In recent years, many learning-based methods \cite{huang2018deepmvs, wang2018mvdepthnet, yao2018mvsnet, im2018dpsnet, kusupati2020normal, yang2020cost, duzceker2021deepvideomvs, long2021multi} have emerged for multi-view depth estimation. They make a promising improvement in depth estimation due to more robust correspondences based on features extracted by deep neural networks (DNNs) \cite{hirschmuller2007stereo, hosni2012fast}.
These methods mainly follow the paradigm of cost volume construction and regularization for depth estimation, as Fig.~\ref{fig:intro} (a) shows.
However, this pipeline still has shortcomings in the way of establishing and utilizing the correspondence. First, the pixel correspondence is predefined by the discrete depth hypotheses. Thus, the predicted depth is sensitive to the sampling strategy of the hypotheses and needs to be modified for different types of scenes. Second, the established correspondence only considers the sparse and fixed pixels from projection and ignores the relation from neighboring areas, which may lead to false matching in textureless or repeated pattern areas. Third, the cost volume which represents the correspondence is processed implicitly by DNNs to fit the depth prediction directly. It is more inclined to overfit the training data compared to the traditional iterative optimization based on dynamically established correspondence.


In essence, multi-view depth estimation is an optimization problem minimizing the pixel matching cost based on stereo correspondence. Traditional methods~\cite{furukawa2015multi} have built an interpretable optimization pipeline, while the powerful representation ability of deep learning can make up for the deficiency of handcrafted correspondence expression. Therefore, we follow the traditional optimization-based framework but apply DNNs for image feature extraction and iterative optimization. Based on the distinctive features extracted by DNNs, dense and reliable correspondences can be established. 
With the help of the approximation ability of DNNs, depth can be better optimized by combining correspondence cue and other regularization terms such as context information. Moreover, the idea of utilizing deep learning to facilitate the correspondence representation and optimization has been verified in optical flow estimation~\cite{teed2020raft} and shows superior performance in various scenarios compared to other methods.

Intuitively, considering multi-view depth estimation as a correspondence-based optimization problem, we propose a novel framework that iteratively optimizes the depth prediction based on projected correlations, as Fig.~\ref{fig:intro} (b) shows.  
We can evaluate the quality of a depth prediction from the correlation volume by reprojecting the point into multiple views according to its estimated depth and checking the correlation between corresponding points.  
The retrieved correlations in turn facilitate the incremental depth refinement. 
Specifically, our framework is composed of three modules: 1) The correlation construction module decouples the pixel correspondences from predefined depth hypotheses and describes all-to-all pixel similarity in an efficient way. It helps to alleviate mismatching on confusing areas by provide neighboring dense correspondences for fusion in the refinement module. 
2) In order to initialize the reasonable depth for further optimization and ensure its generalization ability, the depth initialization module leverages the correlation volumes to estimate the optical flow for predicting an initial depth by triangulation.
3) Correlation-guided depth refinement module mimics the steps of traditional optimization to iteratively update the depth prediction according to the correlations. We design a correlation fusion module and depth updating module to take full use of multi-view correlations in each iteration.
The fusion of correlations is realized by explicitly projecting each point into multiple views according to its predicted depth.
The depth updating module predicts the depth residual according to the fused correlation features, which describe the quality of the correspondence from the estimated depth. Additionally, to handle the problem of occlusion or non-overlapping, which cannot be solved by correspondence, we introduce a learnable context feature as a regularization term to the updating module.
Taking efficiency into account, the depth is initialized and iteratively updated at a low resolution, and finally upsampled to the original resolution in a coarse-to-fine manner to recover details.

In summary, we make the following contributions for multi-view depth estimation in this paper:
\begin{itemize}
   \item Distinct from previous deep multi-view depth estimation methods that rely on predefined depth hypotheses to build correspondence as cost volume and implicitly predict the depth from cost volume regularization, we are the first to explore pixel correlations to build a novel framework that facilitates multi-view depth estimation.

    \item We design a correlation-guided depth refinement module to exploit correspondences for depth prediction iteratively. 
    An effective correlation fusion strategy is proposed to dynamically assess the quality of the current correspondences from projection and guide depth updating. 

    \item Our correlation-based multi-view depth estimation framework is effective and generalized to various scenarios. It produces more precise depth prediction as well as higher visual quality. We demonstrate our new state-of-the-art performance on ScanNet~\cite{dai2017scannet}, DeMoN~\cite{ummenhofer2017demon}, and the best generalization ability on ETH3D~\cite{schops2017multi} and 7Scenes~\cite{shotton2013scene}.
\end{itemize}



\section{Related Work}
Multi-view depth estimation is a popular solution for multi-view stereo (MVS), which focuses on recovering the 3D geometry of the scene from the pixel correspondences among the views. In this section, we will revisit the traditional MVS methods first and then summarize the current learning-based multi-view depth estimation methods. Since we design a novel multi-view depth estimation framework from the perspective of the representation and usage of correspondence, we also introduce the research of correspondence in the field of depth estimation and 3D reconstruction.

\noindent{\textbf{Traditional Multi-view Stereo.}}
Traditional MVS approaches can be summarized into three categories, including point-based\cite{furukawa2009accurate, lhuillier2005quasi}, voxel-based \cite{kar2017learning, ji2017surfacenet, seitz1999photorealistic, kutulakos2000theory}, and depth-based\cite{campbell2008using, lhuillier2005quasi, stuhmer2010real, tola2012efficient, pizzoli2014remode, galliani2015massively, schonberger2016pixelwise, yao2017relative}. 
The point-based and voxel-based methods directly make predictions in 3D space with high consumption of computation and memory. In contrast, depth-based methods are more flexible and efficient by describing the 3D structure through a single scalar for each point. They establish the correspondence between images from multiple views by plane sweep\cite{collins1996space} or patch match \cite{barnes2009patchmatch} and then estimate the depth values by minimizing a global matching cost. 
Furthermore, the estimated depth map can be converted to implicit fields for reconstructing point clouds \cite{merrell2007real} or 3D voxels \cite{newcombe2011kinectfusion}. 
In this paper, we further explore the depth-based MVS by cooperating DNNs. The focus of our work is per-view depth estimation. The final reconstruction can be achieved by integrating any depth fusion methods \cite{zeng20173dmatch, izadi2011kinectfusion}.

\noindent{\textbf{Learning-based Multi-view Depth Estimation.}}
In recent years, learning-based multi-view depth estimation \cite{huang2018deepmvs, wang2018mvdepthnet, yao2018mvsnet, im2018dpsnet, hou2019multi, luo2019p, xue2019mvscrf, yao2019recurrent, sinha2020deltas, gu2020cascade, kusupati2020normal, yang2020cost, duzceker2021deepvideomvs, long2021multi, ke2020deep, yan2020dense} have made great progress. 
DeepMVS~\cite{huang2018deepmvs} builds a set of cost volumes by plane sweep~\cite{collins1996space} from an arbitrary
number of image patches and then aggregates these volumes with a ConvNet to infer depth maps. However, the global information is lost for DeepMVS because it pre-processes the image as patches.
For better learning the global information, MVSNet~\cite{yao2018mvsnet} and DPSNet~\cite{im2018dpsnet} directly extract the features from the whole image and construct the cost volume in the feature space. Their common idea is to build a matching cost volume among views based on a set of depth hypotheses and then regularize the cost volume by 3D convolutions to predict the depth map. 

To further improve the performance, R-MVSNet\cite{yao2019recurrent} reduces the memory consumption by replacing 3D convolution with gated recurrent units (GRU) to sequentially regularize the 2D cost maps along the depth direction. NAS \cite{kusupati2020normal} improves depth estimation by joint learning of normal maps. DeepVideoMVS \cite{duzceker2021deepvideomvs} and EST \cite{long2021multi} extend MVS to videos by fusing temporal information based on long short-term memory or Transformer. However, constructing a cost volume requires a set of predefined discrete depth hypotheses, which inevitably harms the precision and generalization ability. In contrast, we directly make depth predictions based on multi-view pixel-wise correlations rather than predefined depth hypotheses. 

An exception of this cost-volume-based framework is DELTAS~\cite{sinha2020deltas}. To avoid the high memory consumption in cost volume regularization, it first predicts the depths for sparse keypoints by triangulation and then densifies the depth map using 2D convolutions. 
Inspired by their two-step framework, we propose an iterative depth refinement framework and build dense point-wise correspondences instead of sparse correspondences to fully exploit the multi-view correlations.

\noindent{\textbf{Image Correspondence for 3D Reconstruction.}} 
Establishing point correspondences among multi-view images is a fundamental step for 3D reconstruction.
It is common to build sparse correspondences in structure from motion (SfM) \cite{schonberger2016structure, lindenberger2021pixel} and visual simultaneous localization and mapping (SLAM) \cite{mur2017orb, mur2015orb}. The feature point detectors and descriptors can be manually designed \cite{lowe2004distinctive, rublee2011orb} or trained \cite{detone2018superpoint, liu2020extremely} to establish the sparse correspondences. Besides, the optical flow can also be converted as pixel correspondences to provide photometric restriction for SLAM. In comparison, MVS systems typically build dense correspondences based on handcrafted features or learned features by DNNs. 
Cost volume and correlation volume are the two popular representations of correspondences. 
Cost volumes represent the pixel correspondences by plane sweep based on a set of depth hypotheses. 
In comparison, correlation volumes directly depict the pixel-wise similarity between two images and are widely used for stereo matching \cite{yang2008stereo, kanade1994stereo} and optical flow estimation \cite{dosovitskiy2015flownet, teed2020raft}. 
In this paper, we employ correlation volumes to depict pixel correspondences and fully exploit the dense point correlations for accurate depth estimation.

\begin{figure*}
  \centering
  \includegraphics[width=\linewidth]{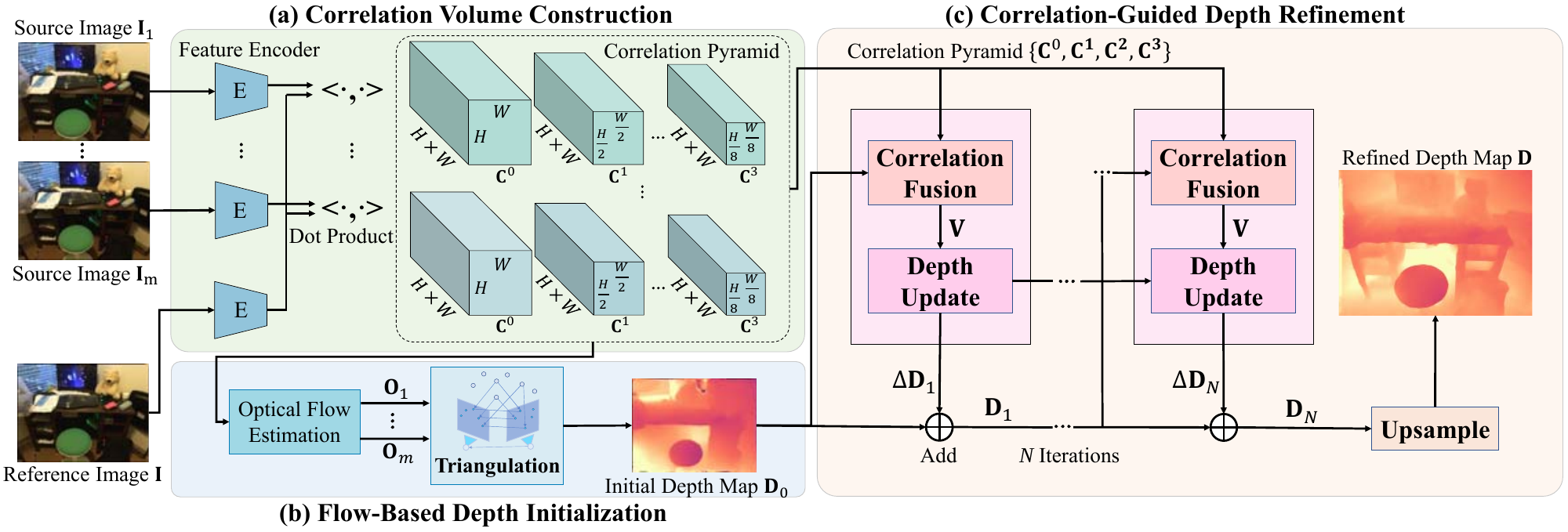}
  \caption{Overview of our correlation-based multi-view depth estimation framework. To estimate the depth map for the reference image $\mathbf{I}$ with other source images $\{\mathbf{I}_{k}\}_{k=1}^{m}$, we first calculate the multi-resolution correlation volumes between the reference image and source images according to encoded image features (a). Second, we calculate an initial depth map at a low resolution by triangulation based on optical flows (b). The initial depth map is iteratively updated based on the point correlations extracted by our fusion module and finally upsampled to the original resolution (c). Our framework is flexible for arbitrary numbers of source images.}
  \label{fig:framework}
\end{figure*}

\section{Our Method}
\subsection{Overview}
Given a series of images including one image $\mathbf{I}$ under a reference view and other source images $\{\mathbf{I}_k\}^{m}_{k=1}$ with its camera intrinsic matrix $\mathbf{K}$ and camera poses $\{\mathbf{T}, \mathbf{T}_{1}, \ldots, \mathbf{T}_{m}\}$, the goal of multi-view depth estimation is to predict a depth map $\mathbf{D}$ under the reference view.
%
Instead of building cost volumes from a set of depth hypotheses, we build correlation volumes between images for iterative depth estimation.
Fig.~\ref{fig:framework} shows the framework of our method, which consists of three parts:

\begin{enumerate}[(a)]
\item \textbf{Correlation volume construction.} To depict the dense point-wise correspondences
between the reference view and other source views, we build a pyramid of correlation volumes based on the encoded features. The correlation pyramid avoids predefined depth hypotheses and represents multi-scale dense correspondence with neighboring relationships to alleviate mismatching in confusing areas.

\item \textbf{Flow-based depth initialization.}
Based on the correlation pyramid, we estimate optical flows between the reference view and other views to establish dense point correspondences. Then we predict an initial depth map by multi-view triangulation according to the estimated dense correspondences. This module can be generalized to various scenes and provides a reasonable initial point for subsequent refinement.
  
\item \textbf{Correlation-guided depth refinement.} 
After initialization, a recurrent unit consisting of correlation fusion and depth updating is designed to refine the depth values progressively considering the point correlations based on the depth values predicted at the last iteration. This design dynamically establishes the multi-view correspondence to reflect the quality of the currently estimated depth and heuristically guides depth updating. Considering efficiency, the entire process is carried out in low resolution and finally upsampled in a coarse-to-fine manner.  
%
\end{enumerate}

\vspace{-0.1cm}
\subsection{Correlation Volume Construction}
\label{sec:correlation volume}

We extract the correlation volume pyramid to depict the pixelwise similarities between the reference view and other source views. For the reference image $\mathbf{I}$ and a source image $\mathbf{I}_{k}$, their image feature maps $\mathbf{E}=f_\theta(\mathbf{I})$ and $\mathbf{E}_k=f_\theta(\mathbf{I}_k)$ are first extracted by a feature encoder $f_{\theta}$.
Denote $H$ and $W$ as the height and width of the feature maps, which are one-eighth of the image resolution.
A 4D correlation volume $\mathbf{C}(\mathbf{I},\mathbf{I}_k)\in \mathbb{R}^{H\times W\times H\times W}$ is calculated by taking the dot product between the feature vectors of a point $\mathbf{p}=(x,y)^T$ in the reference image and a point $\mathbf{q}=(u,v)^T$ in the source image as
\begin{equation} \label{eq:pixel-correlation}
C_{(\mathbf{p},\mathbf{q})}=\mathbf{E}(\mathbf{p})\cdot \mathbf{E}_k(\mathbf{q}).
\end{equation}

In order to represent the point correlations at different scales for collecting sufficient neighboring information, we construct a 4-layer correlation pyramid $\{\mathbf{C}^l|l=0, \dots, 3\}$ by pooling the first correlation volume on the last two dimensions. 
For each layer, the correlation volume $\mathbf{C}^l\in \mathbb R^{H \times W \times \frac{H}{2^l} \times \frac{W}{2^l}}$. 

\vspace{-0.1cm}
\subsection{Flow-based Depth Initialization \label{sec:triangulation}}
According to multi-view geometry, the depth of a point can be calculated by triangulation based on multi-view point correspondences. 
We first estimate the optical flow $\mathbf{O}_{k}$ between the reference image and each source image using RAFT \cite{teed2020raft} to represent the correspondences. Then we compute an initial depth for each pixel by multi-view triangulation as follows.

With the known correspondence and relative pose $\{\mathbf{R}_{k}, \mathbf{t}_{k}\}$ of the reference image to the source images, the triangulation computes a depth $d$ for a pixel $\mathbf{p}$ in the reference image to minimize the projection errors from the reference view to all source images:
\begin{equation}
E_{proj}= \sum_{k}\left\|\left(\mathbf{K}^{-1}\overline{\mathbf{p}}_{k} \right) \times\left( \mathbf{R}_{k} \mathbf{K}^{-1}\overline{\mathbf{p}} d+ \mathbf{t}_{k}\right)\right\|^{2}.
\end{equation}
The relative pose $[\mathbf{R}_{k} \  \mathbf{t}_{k}]$ can be derived from camera poses $\mathbf{T}$ and $\mathbf{T}_{k}$. $\overline{\mathbf{p}}=(x,y,1)^{T}$ is the homogeneous coordinate for a pixel located at $(x,y)$ in the reference image. $\overline{\mathbf{p}}_{k}$ is the homogeneous coordinate of its corresponding point $(u_k,v_k)$ in the source image $\mathbf{I}_k$.  

\begin{figure}
  \centering
  \includegraphics[width=\linewidth]{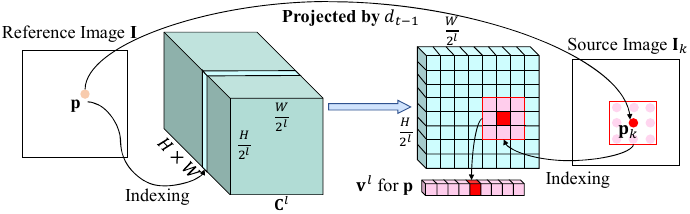}
  \caption{To extract the correlation vector $\mathbf{v}^l$ for $\mathbf{p}$, its corresponding pixel $\mathbf{p}_k$ in source image $\mathbf{I}_k$ is first projected by the current depth $d_{t-1}$. Then $\mathbf{p}$ and the local neighborhood of $\mathbf{p}_k$ are used to index the correlations in $\mathbf{C}^l$ as $\mathbf{v}^l$.}
  \label{fig:fusion}
  
\end{figure}

\vspace{-0.1cm}
\subsection{Correlation-Guided Depth Refinement}
The initial depth prediction is inevitably flawed because of inaccurate optical flow estimation, pure camera rotation, and insufficient resolution for distant points. 
Our key idea is to iteratively revise the depth prediction based on the correlations extracted by the fusion module. 
The fused correlations describe the quality of the correspondences established from the estimated depth and guide the network to update the depth. 
Trading off the accuracy and efficiency, the depth map is initialized and iteratively refined at the low resolution and finally upsampled to the original resolution in a coarse-to-fine manner, as Fig.~\ref{fig:framework} (c) shows.

\subsubsection{Iterative Depth Updating}
\label{sec:lr-depth-update}
We first update the initial depth map iteratively at a low resolution, i.e., $\mathbf{D}_0$ in the resolution of $H \times W$.
In each iteration, we perform correlation fusion via reprojecting each point from the reference view to the source views according to the estimated depth in the last iteration and depth updating based on the fused correlations.

\vspace{0.1cm}
\noindent{\textbf{Correlation Fusion.}} 
At the $t$-th iteration, we have a depth map $\mathbf{D}_{t-1}$ under the reference view estimated at the last iteration.
For each pixel $\mathbf{p}$ in the reference image, we unproject it to the 3D space according to its current depth $d_{t-1}$ and then reproject it to a source view as 
%

\begin{equation}
\label{eq:reprojection}
\homec{p}_k \sim \mathbf{K}\left(\mathbf{R}_{k} \mathbf{K}^{-1} d_{t-1}\homec{p}+\mathbf{t}_{k}\right).
\end{equation}

\dlt{
\begin{equation}
\label{eq:reprojection}
\left(\begin{array}{l}
u \\
v \\
1
\end{array}\right) \sim \mathbf{K}\left[\mathbf{R}_{k} \mathbf{K}^{-1} d_{t-1}\left(\begin{array}{c}
x \\
y \\
1
\end{array}\right)+\mathbf{t}_{k}\right].
\end{equation}
}


 
 As Fig.~\ref{fig:fusion} shown, based on the reprojected position $\mathbf{p}_k$, we fuse the correlations in its local neighborhood $\mathcal{N}(\mathbf{p}_{k})$ which contains all points $\{\mathbf{q}\}$ that $\|\mathbf{q}-\mathbf{p}_k\|_{1} \leq r$, where $r$ is a constant.  
 For each correlation volume $\mathbf{C}^l$ in the correlation pyramid, we take the correlation values $C^{l}_{(\mathbf{p},\mathbf{q})}$ for all points $\mathbf{q}\in \mathcal{N}(\mathbf{p}_{k})$ and concatenate them together with their bilinear-interpolated value and form a correlation vector $\mathbf{v}^{l}$.
 A further concatenation of $\{\mathbf{v}^{l}\}$ from all correlation volumes $\mathbf{C}^l$ forms a fused correlation feature for the point $\mathbf{p}$. 
With similar reprojection and correlation fusion for all the pixels in the reference image to a source image $\mathbf{I}_k$, we can obtain a correlation map $\mathbf{V}_k$. 
We fuse the correlation maps $\{\mathbf{V}_k\}^{m}_{k=1}$ of the $m$ source images into one correlation map $\textbf{V}$ to aggregate the point correlations from multiple views. 
Our default fusion strategy $\mathcal{F}(\cdot)$ is averaging as defined in Eq.~\ref{eql:fusion}. We also compare some other fusion strategies in Sec.~\ref{sec:ablation}. 
\begin{equation}
\label{eql:fusion}
\mathbf{V}=\mathcal{F}\left(\mathbf{V}_{1}, \cdots, \mathbf{V}_{m}\right)=\frac{1}{m}\sum_{k=1}^{m} \mathbf{V}_{k}.
\end{equation}

\noindent{\textbf{Depth Updating.}}
Based on the fused correlation map $\mathbf{V}$ and depth map $\mathbf{D}_{t-1}$ at the previous iteration, we update the depth from a convolutional GRU, as Fig.~\ref{fig:framework} (c) shown. The correlation map $\mathbf{V}$ and depth map $\mathbf{D}_{t-1}$ are first separately passed through two $3 \times 3$ convolutional layers and then concatenated together in the channel dimension as $\mathbf{H}$ before being inputted into the GRU.
However, the correlation map models multi-view correspondence, which can not totally solve the problem from non-overlapping or occlusion. As with traditional optimization methods adding some priors as regularization terms, we addtionally integrate a learnable context feature $\mathbf{F}^{3}$ extracted from the reference image with $\mathbf{H}$ to better handle the non-overlapping or occlusion regions. The details of context feature $\mathbf{F}^{3}$ are described in the following depth upsampling section. The specific architecture of GRU is the same as the one used in RAFT~\cite{teed2020raft}.
A residual depth map $\triangle \mathbf{D}_{t}$ is predicted and added to $\mathbf{D}_{t-1}$.
We start from the initial depth map $\mathbf{D}_0$ at $\frac{1}{8}$ resolution and obtain the depth map $\mathbf{D}_{N}$ at the same resolution after $N$ iterations.

\begin{figure}
  \centering
  \includegraphics[width=\linewidth]{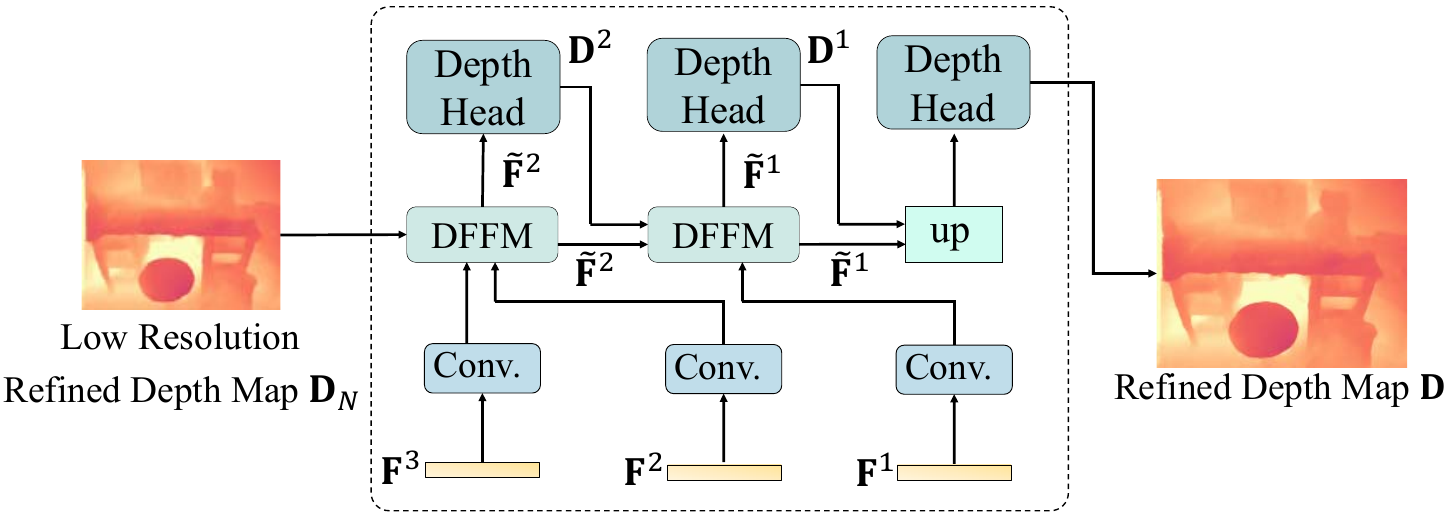}
  \caption{We progressively produce a higher-resolution depth map by fusing the low-resolution depth map and multi-layer context features. The context features are extracted from a lightweight encoder $g$.}
  \label{fig:upsample}
  \vspace{-0.3cm}
\end{figure}

\begin{table*}\normalsize
  \centering
  \caption{Comparison on ScanNet and SUN3D dataset. 
  Our method outperforms other methods by a large margin. (* indicates the dataset is not used during training. EST requires multiple frames to extract temporal information on the video, thus can not be directly tested on SUN3D which only provides two-view image pairs.)}
    \newcolumntype{"}{@{\hskip\tabcolsep\vrule width 1.2pt\hskip\tabcolsep}}
  \label{tab:indoorcomparison}
  \begin{tabular}{@{}l|ccccc"ccccc@{}}
    \toprule
    & & & ScanNet & & & & & SUN3D* \\
    Method & Abs Rel & Abs & Sq Rel & RMSE & $\delta<1.25$ & Abs Rel & Abs & Sq Rel & RMSE & $\delta<1.25$  \\
  \hline
    MVDepth \cite{wang2018mvdepthnet} & 0.1167 & 0.2301 & 0.0596 & 0.3236 & 84.53 & 0.1377 & 0.3199 & 0.1564 & 0.4523 & 82.45\\
    MVDepth-FT & 0.1116 & 0.2087 & 0.0763 & 0.3143 & 88.04 & 0.3092 & 0.7209 & 4.4899 & 1.7180 & 78.73 \\
    DPS \cite{im2018dpsnet} & 0.1200 & 0.2104 & 0.0688 & 0.3139 & 86.40 & 0.1469 & 0.3355 & 0.1165 & 0.4489 & 78.12\\
    DPS-FT & 0.0986 & 0.1998 & 0.0459 & 0.2840 & 88.80 & 0.1274 & 0.2858 & 0.0855 & 0.3815 & 83.96\\
    NAS \cite{kusupati2020normal} & 0.0941 & 0.1928 & 0.0417 & 0.2703 & 90.09 & 0.1271
    & 0.2879 & 0.0852 & 0.3775 & 82.95\\
    DELTAS \cite{sinha2020deltas} & 0.0915 & 0.1710 & 0.0327 & 0.2390 & 91.47 & 0.1245 & 0.2662 & 0.0741 & 0.3602 & 85.51\\
    EST \cite{long2021multi} & 0.0812 & 0.1505 & 0.0298 & 0.2199 & 93.13 & - & - & - & - & -\\
    Ours & \textbf{0.0607} & \textbf{0.1162} & \textbf{0.0205} & \textbf{0.1915} & \textbf{95.99} & \textbf{0.1121} & \textbf{0.2552} & \textbf{0.0661} & \textbf{0.3369} & \textbf{87.15}\\
    \bottomrule
  \end{tabular}
\end{table*}

\subsubsection{Coarse-to-Fine Upsampling}
\label{sec:depth-upsample}
The construction of correlation volumes and the iterative depth updating are performed at $\frac{1}{8}$ resolution for efficiency.
Then we design a depth upsampling module to recover object details in a coarse-to-fine manner by three $2\times$ upsampling layers, as Fig.~\ref{fig:upsample} shows.

At each upsampling layer, in order to complement geometric details and solve the ambiguity of low-texture regions, we integrate the context features extracted from the reference image with the low-resolution depth maps. 
More specifically, we extract three-scale context features $\{\mathbf{F}^{1}, \mathbf{F}^{2}, \mathbf{F}^{3}\}$ from the reference image $\mathbf{I}$ using a context encoder $g$, which has the same architecture as the feature encoder $\theta$. 
In the depth and feature fusion module (DFFM), we first upsample the depth map $\vb{D}^{l+1}$ and the context feature map ($\mathbf{F}^{l+1}$ when $l=2$) or fused feature ($\Tilde{\mathbf{F}}^{l+1}$ for $l=1,0$), and fuse them with the context feature $\mathbf{F}^{l}$ at a higher resolution to generate a fused feature map $\Tilde{\mathbf{F}}^{l}$ in $2\times$ resolution.
Based on the fused feature map, we predict a depth map $\mathbf{D}^{l}$ with a depth head that consists of two convolutional layers. 
Implementation details of the DFFM and depth head are described in the supplementary material. 
At the last layer, we obtain the final depth map $\vb{D}$ in the resolution same as the input image.

\vspace{-0.2cm}
\subsection{Loss Function}
\label{sec:loss}
The network is supervised with the $L_1$ loss between ground truth and predicted depth at both low resolution and original resolution.
Specifically, the loss function is
\begin{equation}\label{eql:loss}
\mathcal{L}=\sum_{t=1}^{N} \gamma^{t-N}\left\|\mathbf{D}_{g t}^{low}-\mathbf{D}_{t}^{low}\right\|_{1} + \|\mathbf{D}_{gt} - \mathbf{D}\|_{1},
\end{equation}
where $N$ is the total number of iterations. $\gamma$ is a constant and set to $0.8$ in our experiments. $\mathbf{D}_{gt}^{low}$ is the ground truth depth map in $\frac{1}{8}$ resolution and $\mathbf{D}_{t}^{low}$ is the the estimated depth map at $t$-th iteration in $\frac{1}{8}$ resolution. $\mathbf{D}_{gt}$ is the ground truth depth at the original resolution and $\mathbf{D}$ is the final depth map obtained by the depth upsampling module.
\section{Experiments}
\begin{table*}[t!]\normalsize
  \newcommand{\tabincell}[2]{\begin{tabular}{@{}#1@{}}#2\end{tabular}}
  \caption{Evaluation on DeMoN and ETH3D. * indicates the datasets are not used during training. Hybrid means containing both indoor and outdoor scenes. Bold indicates the best and underline indicates the second.}
  \label{tab:demoncompare}
  \newcolumntype{"}{@{\hskip\tabcolsep\vrule width 1.2pt\hskip\tabcolsep}}
  \centering
  \resizebox{.9\textwidth}{!}{%
  \setlength{\tabcolsep}{1.0mm}{
  \begin{tabular}{@{}l|lr|lr|lr"lr|lr@{}}
    \toprule
    \multirow{2}{*}{Method} & \multicolumn{2}{c|}{SUN3D (Indoor)} & \multicolumn{2}{c|}{RGBD (Indoor)} & \multicolumn{2}{c"}{Scenes11 (Synthetic)} & \multicolumn{2}{c|}{MVS* (Outdoor)} & \multicolumn{2}{c}{ETH3D* (Hybrid)}\\
    & AbsRel & $\delta<1.25$ & AbsRel & $\delta<1.25$ & AbsRel & $\delta<1.25$ & AbsRel & $\delta<1.25$ & AbsRel & $\delta<1.25$ \\
    
    \midrule
    COLMAP~\cite{schonberger2016structure} & 0.6232 & 32.66
                                        & 0.5389 & 27.49
                                        & 0.6249 & 38.97
                                        & 0.3841 & 48.19
                                        & 0.324 & 86.5 \\
    DeMoN~\cite{ummenhofer2017demon}~   & 0.2137 & 73.32
                                        & 0.1569 & 80.11
                                        & 0.5560 & 49.63
                                        & 0.3105 & 64.11
                                        & 0.191 & 73.3\\
    DeepMVS~\cite{huang2018deepmvs}     & 0.2816 & 56.22
                                        & 0.2938 & 54.93
                                        & 0.2100 & 68.81
                                        & 0.2305 & 67.37
                                        & 0.178 & 85.8\\
    DPSNet~\cite{im2018dpsnet}          & 0.1469 & 78.12
                                        & 0.1508 & 80.41
                                        & 0.0500 & 96.14
                                        & 0.0813 & 88.53
                                        & 0.099 & 86.3\\
    NAS~\cite{kusupati2020normal}       & \underline{0.1271} & \underline{82.95}
                                        & \underline{0.1314} & \underline{85.65}
                                        & \textbf{0.0380} & \underline{97.54}
                                        &  \underline{0.0679} & \underline{90.54}
                                        & \textbf{0.090} & \textbf{88.6} \\
    Ours & \textbf{0.1068} & \textbf{87.56} & 
           \textbf{0.0943} & \textbf{90.95} & \underline{0.0388} & \textbf{97.63} &
           \textbf{0.0656} & \textbf{90.64} &
           \underline{0.091} & \underline{88.2} \\
  \midrule
  \end  
  {tabular}}}
\end{table*}

\subsection{Dataset and Metrics}
\subsubsection{Datasets}
Deep multi-view depth estimation methods choose different benchmarks for evaluation. DeepMVS~\cite{huang2018deepmvs}, DPSNet~\cite{im2018dpsnet}, NAS~\cite{kusupati2020normal} use DeMoN~\cite{ummenhofer2017demon} and ETH3D~\cite{schops2017multi} while DELTAS~\cite{sinha2020deltas}, CNM~\cite{long2020occlusion}, EST~\cite{long2021multi} are evaluated on ScanNet~\cite{dai2017scannet}, SUN3D~\cite{xiao2013sun3d}, and 7Scenes~\cite{shotton2013scene}. In order to fully demonstrate the superiority of our method, we evaluate it on all datasets metioned above. Since we focus on depth estimation rather than reconstruction, the MVS benchmarks DTU~\cite{aanaes2016large} and Tank $\&$ Templates~\cite{Knapitsch2017} are not considered. 

\noindent{\textbf{ScanNet}}~\cite{dai2017scannet} contains more than 1600 indoor scenes with different environments, layouts and textures. Color images, ground-truth (GT) depth maps, camera intrinsics, and extrinsics are provided. We follow the ScanNet v2 official split to divide the training, validation, and test sets. 

\noindent{\textbf{DeMoN}} is a mixed dataset introduced by Ummenhofer et al.~\cite{ummenhofer2017demon} for multi-view depth estimation. Its data comes from MVS~\cite{schonberger2016structure}, SUN3D~\cite{xiao2013sun3d}, RGBD~\cite{sturm2012benchmark}, and Scenes11~\cite{ummenhofer2017demon}. In particular, MVS contains real-world outdoor environments, which is only used for evaluation. Scenes11 is a synthetic dataset generated by random shapes and motions whereas SUN3D and RGBD consist of real word indoor environments.

\noindent{\textbf{ETH3D}}~\cite{schops2017multi} is a large-scale MVS dataset consisting of calibrated high-resolution images of indoor and outdoor challenge scenes with large viewpoint variations. Its test set is used in our paper for the evaluation of methods' generalization ability.


\noindent{\textbf{7Scenes}}~\cite{shotton2013scene} is collected from 7 different indoor scenes. It consists of tracked RGB-D camera frames in which depth and RGB images are not aligned. Thus, we do not conduct a quantitative evaluation on 7Scenes.
Instead, we reconstruct the scenes based on the predicted depth and the implementation of TSDF fusion \cite{zeng20173dmatch}.

\subsubsection{Evaluation Metrics}
We use five standard metrics for quantitative evaluation:

\begin{itemize}
\item Absolute relative depth error (Abs Rel):
\begin{small}
$\frac{1}{M} \sum_{i=1}^{M}\left\|d_{g t}^{i}-d^{i}\right\| / d_{g t}^{i},$
\end{small}

\item Absolute depth error (Abs):
\begin{small}
$\frac{1}{M} \sum_{i=1}^{M}\left\|d_{g t}^{i}-d^{i}\right\|,$
\end{small}

\item Square relative error (Sq Rel):
\begin{small}
$\frac{1}{M} \sum_{\mathrm{i}=1}^{M}\left\|d_{g t}^{i}-d^{i}\right\|^{2} / d_{g t}^{i},$
\end{small}

\item Root mean square error (RMSE):
\begin{small}
$\sqrt{\frac{1}{M} \sum_{i=1}^{M}\left\|d_{g t}^{i}-d^{i}\right\|^{2}},$
\end{small}

\item The inlier ratio with threshold 1.25 ($\delta <$ 1.25):
\begin{small}
$\delta=\max \left(\frac{d_{g t}^{i}}{d^{i}}, \frac{d^{i}}{d_{g t}^{i}}\right)<1.25,$
\end{small}
\end{itemize}
\noindent{where $M$ is the number of pixels that are valid in the depth ground truth while $d^{i}_{gt}$ and $d^{i}$ are the ground truth and predicted depth values for pixel $i$.}



\begin{figure*}
  \centering
  \includegraphics[width=\linewidth]{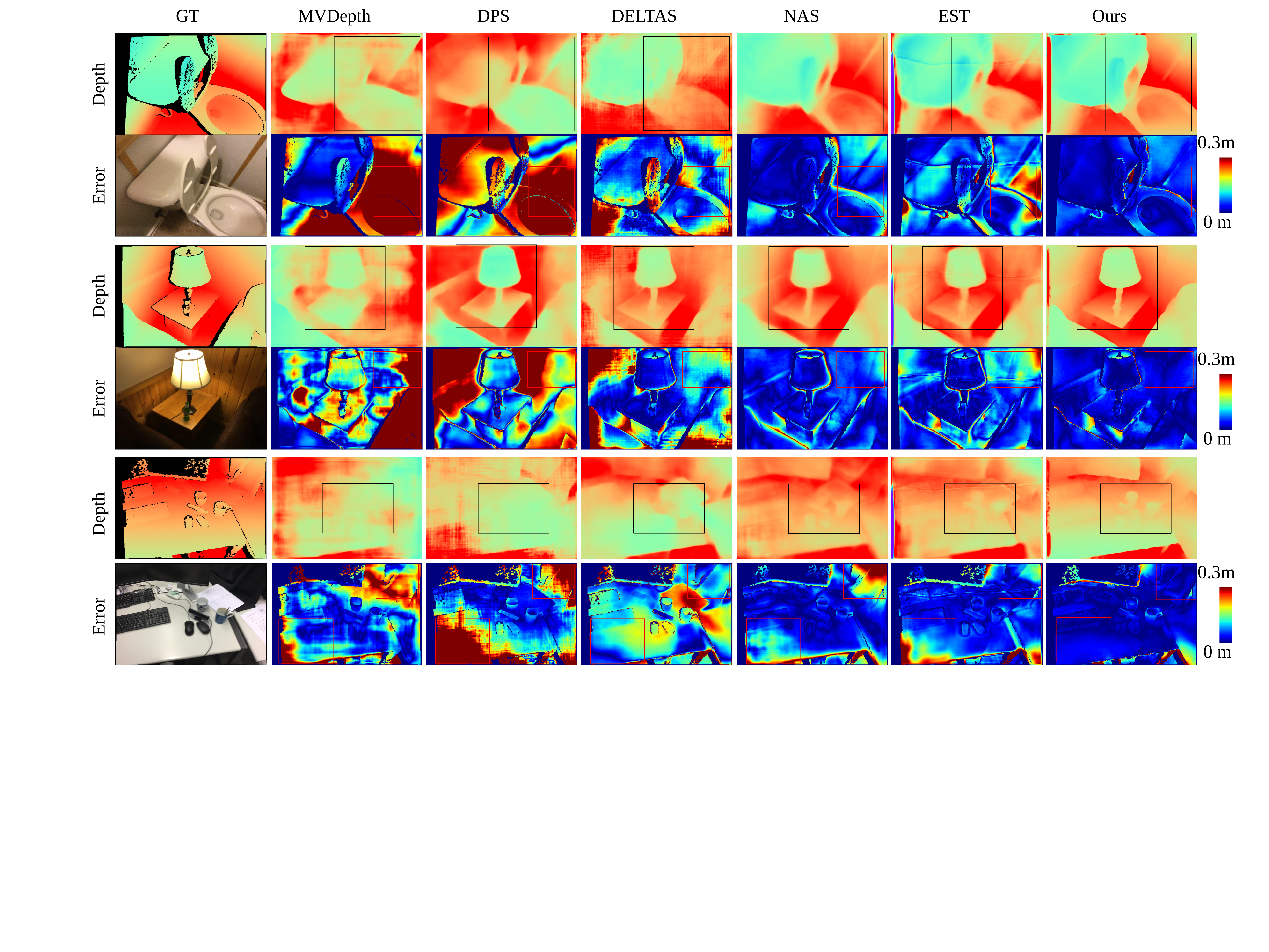}
  \caption{Qualitative comparison with other learning-based methods. Benefiting from our dense and accurate correspondences, our method recovers more object details with sharp boundaries (black box) and generates the most precise depth prediction, especially at textureless regions (red boxes).}
  \label{fig:quatitative_result}
\end{figure*}

\vspace{-0.2cm}
\subsection{Implementation Details}
We implement our model with PyTorch and use the AdamW optimizer \cite{loshchilov2017decoupled}. The learning rate is 6e-4 and we clip gradients to the range $[-1,1]$. Our correlation-guided depth refinement module is initialized from scratch with random weights while the feature encoder $\theta$ and optical flow estimation network is pretrained from RAFT. The threshold $r$ in the correlation fusion module is set to 3. The iteration times on the depth refinement stage is 12 for both training and testing. To evaluate our framework on the different benchmarks with other methods fairly, we train two independent models, one on ScanNet and the other on DeMoN.
We train our model with the batch size of 48 on 8 NVIDIA V100 GPUs for 80k iterations. It takes about 36 hours for training.
On both datasets, the input image resolution is 640 $\times$ 480. For ScanNet, we follow EST's view selection strategy. For each reference image $\mathbf{I}_k$, we choose its four neighboring frames $\{\mathbf{I}_{k-2}, \mathbf{I}_{k-1}, \mathbf{I}_{k+1}, \mathbf{I}_{k+2}\}$ with the interval of ten as the source images. For DeMoN, we follow its official train/test split as previous methods~\cite{im2018dpsnet,kusupati2020normal}. 

\vspace{-0.2cm}
\subsection{Multi-View Depth Estimation Evaluation}

\begin{figure*}
  \caption{TSDF reconstruction of scenes in the ScanNet (the first two rows) and 7Scenes (the last two rows) dataset. Our approach better recovers the shape details, such as the cabinet and door (red box). Our depth prediction leads to smoother and denser surfaces with the minimum distortion (green box) and fewer noises (blue box). \textbf{Better viewed when zooming in}.}
  \label{fig:reconstruction}
  \centering
  \includegraphics[width=\linewidth]{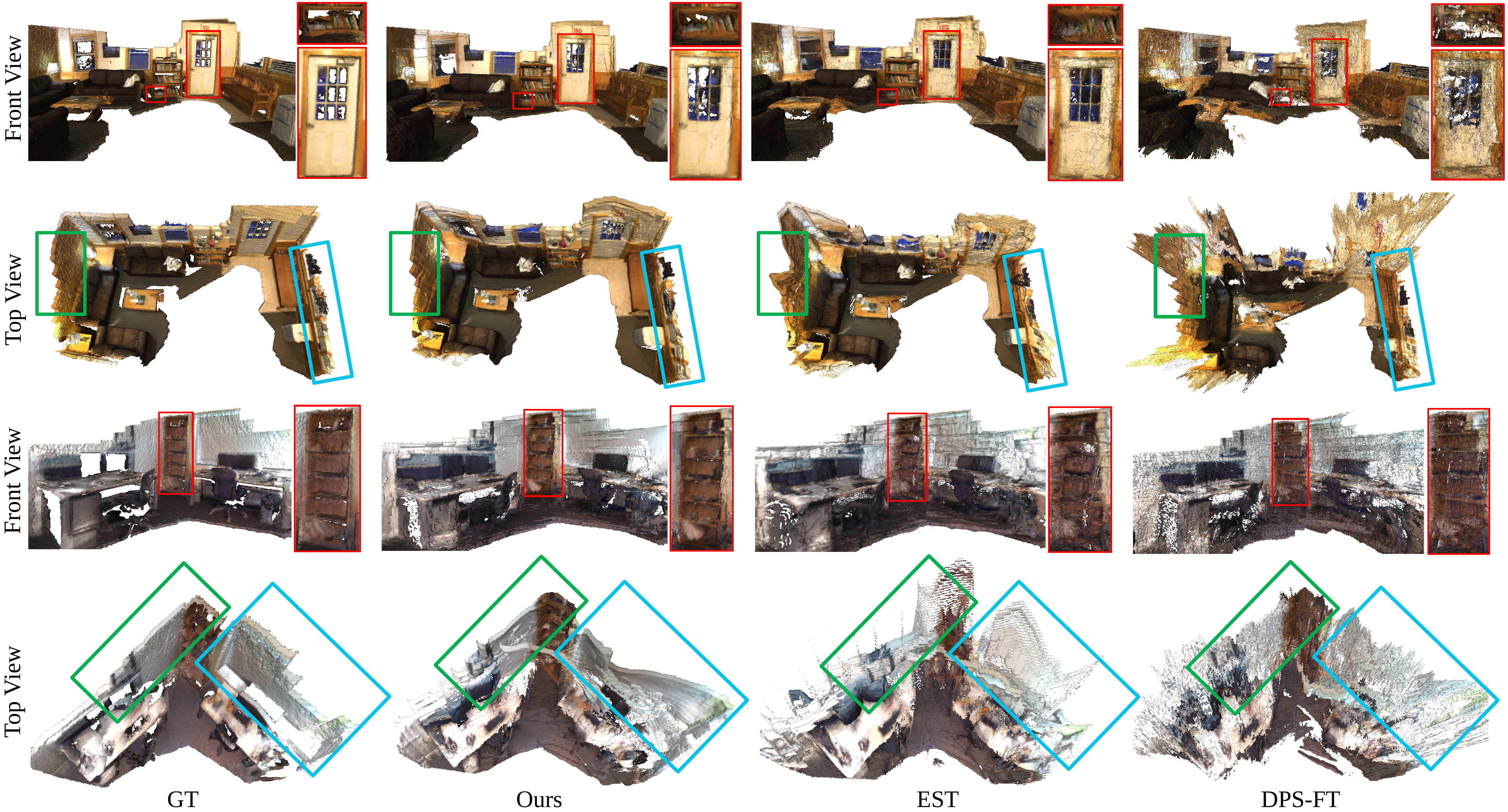}
\end{figure*}

\noindent{\textbf{Quantitative Evaluation.}}
For ScanNet, we compare our method with five learning-based multi-view depth estimation approaches: MVDepthNet \cite{wang2018mvdepthnet}, DPSNet \cite{im2018dpsnet}, NAS \cite{kusupati2020normal}, DELTAS \cite{sinha2020deltas}, and EST \cite{long2021multi}. 
%
Since MVDepth and DPS are originally trained on the DeMoN \cite{ummenhofer2017demon} dataset, for a fair comparison, they are fine-tuned on ScanNet. 
As shown in Table~\ref{tab:indoorcomparison}, our model significantly outperforms other methods over all the metrics on ScanNet datasets. Even the SOTA method EST explores spatial-temporal relations of multiple frames, we still surpass it by a large margin. It shows the superiority of our multi-view correspondence representation compared to the cost volume for depth estimation. Besides, our approach also yields great improvement over DELTAS, which does not build cost volume but establishes sparse correspondence between views. Our method gets better performance by building point-wise dense correspondences.

For DeMoN, we compare with the traditional geometry-based multi-view stereo (COLMAP), depth from unstructured two-view stereo (DeMoN) and deep cost-volume-based methods (DeepMVS, DPSNet, NAS). The results are reported in Table~\ref{tab:demoncompare}. We achieve the best performance overall on both indoor and outdoor datasets, especially a large improvement on indoor datasets. This validates our superiority by avoiding predefined depth ranges since the depth distribution predefined by the previous deep cost-volume-based methods deviates greatly from the indoor scenes for training mixed data on both indoor and outdoor environments.
 


\noindent{\textbf{Qualitative Evaluation.}}
To demonstrate the advantages of our method in recovering global structure and shape details, we also visualize some depth maps and reconstruct indoor scenes for comparison.
As shown in Fig.~\ref{fig:quatitative_result}, benefiting from our dense and accurate pointwise correspondences, our method recovers more details and sharper boundaries and achieves the most accurate depth estimation on the textureless regions. 
As shown in the first two rows in Fig.~\ref{fig:reconstruction}, the scene in ScanNet is constructed using the predicted depth over the whole sequence. Our methods recover smoother surface (door in red box), better details (bookcase in red box), and correct structure with fewer noises (green box and blue box on the top view).
%
\vspace{-0.1cm}
\subsection{Generalization to Unseen Datasets}
To demonstrate our generalization ability, we also compare the methods on unseen datasets. For the model trained on ScanNet, we choose SUN3D and 7Scenes. For the model trained on DeMoN, we evaluate on MVS and ETH3D.

\noindent{\textbf{Quantitative Evaluation.}}
As shown in Table~\ref{tab:indoorcomparison}, even our model is not trained on SUN3D, it still outperforms all the other methods, including those trained on SUN3D, i.e., MVDepth, DPS, and NAS.
EST requires multiple frames as input to extract temporal relations, thus it can not be directly tested on SUN3D which only provides two-view image pairs. 
As shown in Table~\ref{tab:demoncompare}, our method also achieves the best performance on MVS and the comparable results on ETH3D with SOTA NAS. However, our method is more efficient (Fig.~\ref{fig:efficiency}) and does not need additional normal map for supervision compared to NAS. The generalization ability of our framework comes from the general representation and interpretable usage of correspondence.

\noindent{\textbf{Qualitative Evaluation.}}
 As shown in the last two columns in Fig.~\ref{fig:reconstruction}, we reconstruct an office scene in unseen dataset 7Scenes by $50$ images sampled every ten frames. Our approach better recovers the scene structure (top view) and object details (front view).
Besides, the reconstructed surfaces are smoother and contain fewer noises compared with other methods. This further proves the generalization ability of our method and its superiority in reconstruction.

\vspace{-0.2cm}
\subsection{Efficiency Analysis}

\begin{figure}
  \centering
  \includegraphics[width=\linewidth]{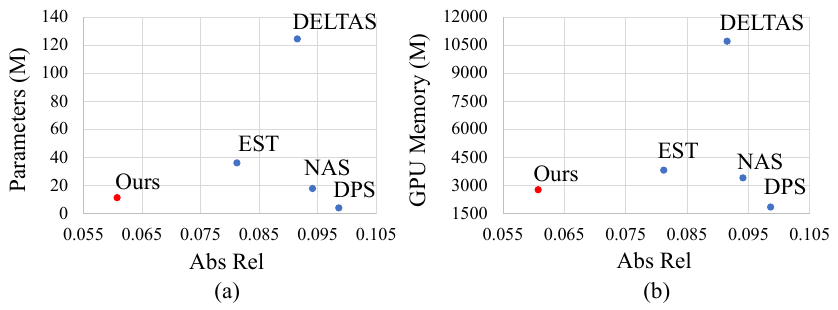}
  \caption{Efficiency comparison with other methods concerning depth prediction errors on ScanNet with model parameters (a) and memory cost (b).}
  \label{fig:efficiency}
\end{figure}

Our correlation-based depth estimation framework not only produces high-quality depth maps, but also costs lower computations and memory.
In Fig.~\ref{fig:efficiency}, we compare the number of network parameters and GPU memory usage of the model with the state-of-the-art learning-based multi-view depth estimation methods~\cite{im2018dpsnet, kusupati2020normal, sinha2020deltas, long2021multi}. 
These methods are tested with the same setting: one reference image with four source images at the resolution of $640 \times 480$ on ScanNet. In comparison, our model is lightweight, which has the second-lowest number of parameters (11.5M) and GPU memory cost (2,778M) while ensuring the highest accuracy. Although DPS \cite{im2018dpsnet} has fewer parameters and memory costs than our model, our method outperforms it on accuracy by a large margin. 

\vspace{-0.2cm}
\subsection{Ablation study\label{sec:ablation}}
We conduct a set of experiments to demonstrate the impact of each component in our framework, including correlation fusion, iteration times, and context features. More ablation studies on the number of source views, depth initialization, and depth upsampling can be found in the supplementary material.

\noindent{\textbf{Correlation Fusion of Neighboring Points.}} 
As described in Sec.~\ref{sec:lr-depth-update}, we first construct a correlation map $\mathbf{V}_k$ between the reference image and a source image by reprojecting each pixel to the source view and concatenating the correlations of neighboring points. 
Another option is directly concatenating the correlations at a pixel in the correlation pyramid in the last two dimensions and applying 2D convolution, similar to the cost volume regularization in previous methods.
As shown in Table~\ref{tab:fusion}, this `Conv. Map' does not perform well compared to our reprojection strategy that establishes a explicit relationship between the depth prediction and correlation volumes.

\noindent{\textbf{Fusion Correlation of Multiple Views.}}
When fusing the correlation maps $\{\mathbf{V}_k\}^m_{k=1}$ with multiple source views, different fusion methods can be used, such as averaging, using their variance, max-pooling, or $1 \times 1$ convolution. 
As Table~\ref{tab:fusion} shows, `Convolution' leads to the best results, showing its strong ability in learning effective fusion patterns. However, the convolution brings more parameters and requires a fixed number of input frames, which is not flexible enough. Similarly, the variance strategy is not suitable for two-view stereos. Contrarily, the averaging and max-pooling operations work for arbitrary view numbers with slight accuracy loss. We finally select `Averaging' because of its comparable results with max-pooling and better generalization ability on other datasets.   

\begin{table}
  \caption{Comparison of different fusion strategies for correlation maps (tested on ScanNet).}
  \label{tab:fusion}
  \centering
  \begin{tabular}{@{}l|cccc@{}}
    \toprule
    Fusion Strategy & Abs Rel & Abs & RMSE & $\sigma <$ 1.25 \\
    \hline
    Conv. Map & 0.0738 & 0.1330 & 0.2018 & 93.75 \\
   \hline
    \underline{Averaging}  & 0.0607 & 0.1162 & 0.1915 & 95.99 \\
    Max-Pooling & 0.0604 & 0.1157 & 0.1909 & 95.97 \\
    Variance & 0.0605 & 0.1150 & 0.1902 & 95.98 \\
    Convolution & \textbf{0.0577} & \textbf{0.1108} & \textbf{0.1855} & \textbf{96.34} \\
    \bottomrule
  \end{tabular}
\end{table} 

\begin{figure}
  \centering
  \includegraphics[width=\linewidth]{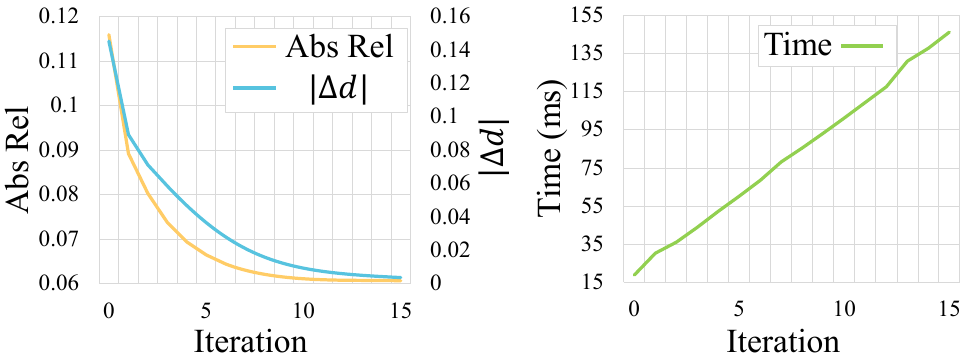}
  \caption{The iterative depth updates, the absolute relative error between the predicted depth and ground truth, and time cost at different iterations.}
  \label{fig:iteration}
\end{figure}

\noindent{\textbf{Iteration Times.}}
For the iterative depth updating described in Sec.~\ref{sec:lr-depth-update}, we verify the convergence of the depth update value $|\Delta d|$ and determine the optimal number of iterations considering the mean $|\Delta d|$ of all training data and the absolute relative error of the test data. 
It shows that our iterative depth updating method progressively refines the initial depth map and converges in a few iterations.
According to the trends of depth updating shown in Fig.~\ref{fig:iteration}, we select $N=12$ iterations in our experiments. In addition, the time of depth refinement increases linearly with the iteration, which reaches about 115ms at the 12th iteration.

\begin{figure}
  \centering
  \includegraphics[width=\linewidth]{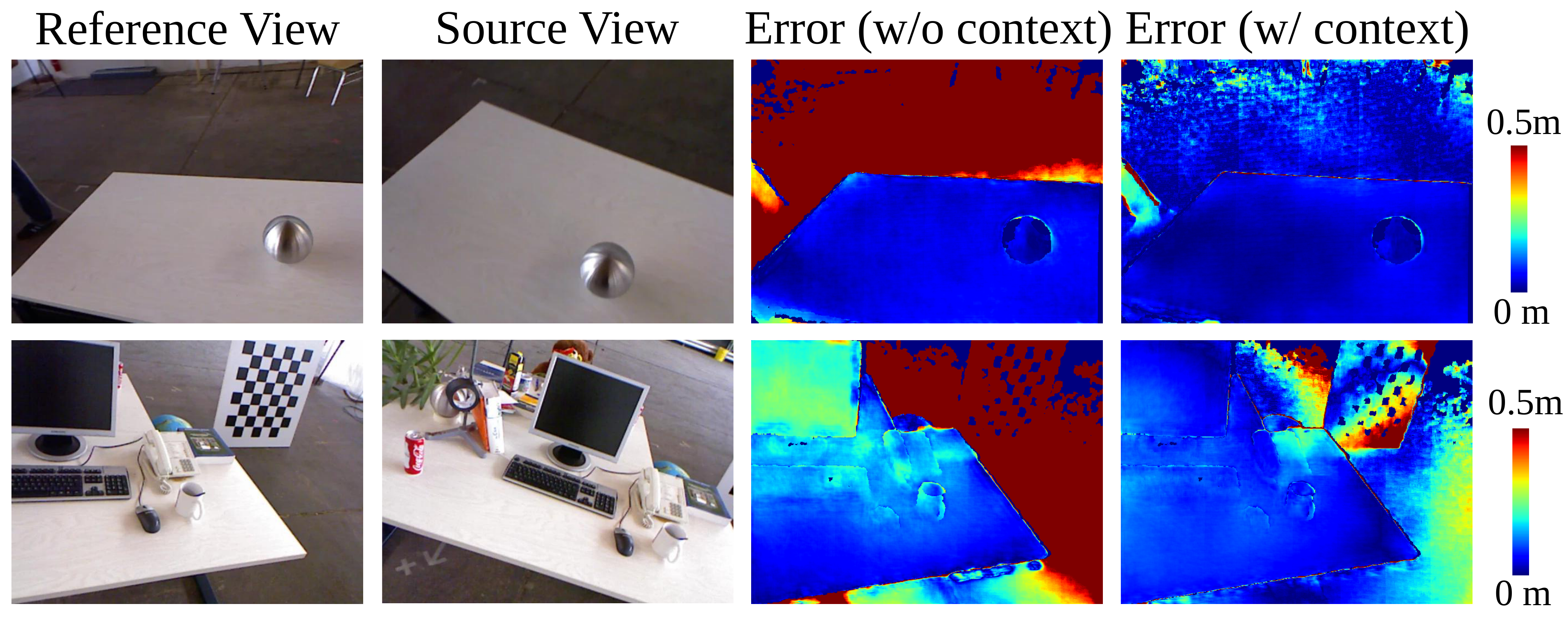}
  \caption{Comparison of the model with or without context features. Context feature provide extra information for the textureless region or the region does not overlap between the reference image and target image.}
  \label{fig:context}
\end{figure}

\noindent{\textbf{Context Features.}}
While the correlation volumes mainly describe pixel-wise similarities between multiple views, the context feature extracted from the reference image provides more spatial context between neighboring points and serves as a regularization term during iterative updating. As shown in Fig.~\ref{fig:context}, it improves the depth estimation on low-texture regions and the non-overlapping regions when the views change greatly.
We compare different implementations of the context encoder $g$ in Table~\ref{tab:ablation study} on ScanNet.
We test a model (`None') without using context features for depth refinement. 
Since the optical flow estimation module in RAFT~\cite{teed2020raft} also employs the context feature, we test another variant of our model 
(`Shared') which directly uses the pretrained context encoder $g$ in RAFT~\cite{teed2020raft} for our depth refinement. 
In our final model, we train an independent context encoder $g$ for depth refinement. 
As shown in Table~\ref{tab:ablation study}, training an independent context encoder for depth refinement (`Indep.') leads to the best performance. However, even without using the context feature, our model `None' still outperforms existing learning-based methods, demonstrating the effectiveness of correlation-based depth estimation.  

\begin{table}
  \caption{Ablation study on context features. We select the underlined settings in our final model.}
  \label{tab:ablation study}
  \centering
  \footnotesize
  \begin{tabular}{@{}cl|cccc@{}}
    \toprule
      & Strategy & Abs Rel & Abs & RMSE & $\sigma <$ 1.25 \\ 
    \hline
    \multirow{3}{*}{Context Feat.} & None & 0.0638 & 0.1220 & 0.2030 & 95.42 \\
    & Shared & 0.0625 & 0.1214 & 0.2028 & 95.47 \\
    & \underline{Indep.} & \textbf{0.0607} & \textbf{0.1162} & \textbf{0.1915} & \textbf{95.99} \\
    \hline
  \end{tabular}
\end{table}

\section{Discussion and Limitation}
The experiment results on different datasets demonstrate the superiority of our model by introducing a new way to represent and exploit pixel correspondences into multi-view depth estimation. Compared to the previous learning-based methods, we take advantage of deep learning and multi-view geometry to endow the model with stronger generalization capability.
Furthermore, since pixel correspondence is a fundamental and generic medium representation, our framework can be extended to many other multi-view or video tasks, such as multi-view pose estimation, video tracking, etc.
It is worth exploring the generality of the correlation volume among these vision tasks. 
Another direction is to explore whether joint learning can bring more robust pixel correspondence.
However, our framework also has limitations. 
Currently, we have not made specific designs for dynamic objects that violate the depth-based reprojection principle. Adding semantic priors or estimating motion flows can be potential solutions for depth prediction in dynamic scenes.


%
 
\section{Conclusion}
In this paper, we propose a novel multi-view depth estimation framework that fully exploits multi-view correlations to mimic traditional optimization process. 
Distinctive to cost-volume-based methods that require a set of predefined depth hypotheses, we directly infer dense point-wise correspondence for depth estimation from multi-view correlations.
We design an correlation-guided depth refinement module that incrementally updates the depth prediction from image correlations. 
Our correlation fusion based on point reprojection and depth updating based on fused correlations effectively integrates multi-view geometry for structure recovery and DNNs for correlation fusion.
Besides, the multi-scale correlations can alleviate mismatching on confusing areas by fusing sufficient neighboring information. 
The combination of DNNs and multi-view geometry endows our framework with stronger inferring ability and generalization capability. 
Experiments on ScanNet, DeMoN, ETH3D, and 7Scenes demonstrate that our model achieves state-of-the-art performance and is well generalized to various scenes. 



\bibliographystyle{IEEEtran}
\bibliography{egbib}

\begin{thebibliography}{10}
\providecommand{\url}[1]{#1}
\csname url@samestyle\endcsname
\providecommand{\newblock}{\relax}
\providecommand{\bibinfo}[2]{#2}
\providecommand{\BIBentrySTDinterwordspacing}{\spaceskip=0pt\relax}
\providecommand{\BIBentryALTinterwordstretchfactor}{4}
\providecommand{\BIBentryALTinterwordspacing}{\spaceskip=\fontdimen2\font plus
\BIBentryALTinterwordstretchfactor\fontdimen3\font minus
  \fontdimen4\font\relax}
\providecommand{\BIBforeignlanguage}[2]{{%
\expandafter\ifx\csname l@#1\endcsname\relax
\typeout{** WARNING: IEEEtran.bst: No hyphenation pattern has been}%
\typeout{** loaded for the language `#1'. Using the pattern for}%
\typeout{** the default language instead.}%
\else
\language=\csname l@#1\endcsname
\fi
#2}}
\providecommand{\BIBdecl}{\relax}
\BIBdecl

\bibitem{schonberger2016pixelwise}
J.~L. Sch{\"o}nberger, E.~Zheng, J.-M. Frahm, and M.~Pollefeys, ``Pixelwise
  view selection for unstructured multi-view stereo,'' in \emph{European
  Conference on Computer Vision}.\hskip 1em plus 0.5em minus 0.4em\relax
  Springer, 2016, pp. 501--518.

\bibitem{song2014sliding}
S.~Song and J.~Xiao, ``Sliding shapes for {3D} object detection in depth
  images,'' in \emph{European Conference on Computer Vision}.\hskip 1em plus
  0.5em minus 0.4em\relax Springer, 2014, pp. 634--651.

\bibitem{couprie2013indoor}
C.~Couprie, C.~Farabet, L.~Najman, and Y.~Lecun, ``Indoor semantic segmentation
  using depth information,'' in \emph{First International Conference on
  Learning Representations (ICLR 2013)}, 2013, pp. 1--8.

\bibitem{campbell2008using}
N.~D. Campbell, G.~Vogiatzis, C.~Hern{\'a}ndez, and R.~Cipolla, ``Using
  multiple hypotheses to improve depth-maps for multi-view stereo,'' in
  \emph{European Conference on Computer Vision}.\hskip 1em plus 0.5em minus
  0.4em\relax Springer, 2008, pp. 766--779.

\bibitem{collins1996space}
R.~T. Collins, ``A space-sweep approach to true multi-image matching,'' in
  \emph{Proceedings CVPR IEEE Computer Society Conference on Computer Vision
  and Pattern Recognition}.\hskip 1em plus 0.5em minus 0.4em\relax IEEE, 1996,
  pp. 358--363.

\bibitem{hosni2012fast}
A.~Hosni, C.~Rhemann, M.~Bleyer, C.~Rother, and M.~Gelautz, ``Fast cost-volume
  filtering for visual correspondence and beyond,'' \emph{IEEE Transactions on
  Pattern Analysis and Machine Intelligence}, vol.~35, no.~2, pp. 504--511,
  2012.

\bibitem{hirschmuller2007stereo}
H.~Hirschmuller, ``Stereo processing by semiglobal matching and mutual
  information,'' \emph{IEEE Transactions on Pattern Analysis and Machine
  Intelligence}, vol.~30, no.~2, pp. 328--341, 2007.

\bibitem{huang2018deepmvs}
P.-H. Huang, K.~Matzen, J.~Kopf, N.~Ahuja, and J.-B. Huang, ``{DeepMVS}:
  Learning multi-view stereopsis,'' in \emph{Proceedings of the IEEE Conference
  on Computer Vision and Pattern Recognition}, 2018, pp. 2821--2830.

\bibitem{wang2018mvdepthnet}
K.~Wang and S.~Shen, ``{MVDepthNet}: Real-time multiview depth estimation
  neural network,'' in \emph{2018 International Conference on 3D Vision
  (3DV)}.\hskip 1em plus 0.5em minus 0.4em\relax IEEE, 2018, pp. 248--257.

\bibitem{yao2018mvsnet}
Y.~Yao, Z.~Luo, S.~Li, T.~Fang, and L.~Quan, ``{MVSNet}: Depth inference for
  unstructured multi-view stereo,'' in \emph{Proceedings of the European
  Conference on Computer Vision (ECCV)}, 2018, pp. 767--783.

\bibitem{im2018dpsnet}
S.~Im, H.-G. Jeon, S.~Lin, and I.~S. Kweon, ``{DPSNet}: End-to-end deep plane
  sweep stereo,'' in \emph{International Conference on Learning
  Representations}, 2018.

\bibitem{kusupati2020normal}
U.~Kusupati, S.~Cheng, R.~Chen, and H.~Su, ``Normal assisted stereo depth
  estimation,'' in \emph{Proceedings of the IEEE/CVF Conference on Computer
  Vision and Pattern Recognition}, 2020, pp. 2189--2199.

\bibitem{yang2020cost}
J.~Yang, W.~Mao, J.~M. Alvarez, and M.~Liu, ``Cost volume pyramid based depth
  inference for multi-view stereo,'' in \emph{Proceedings of the IEEE/CVF
  Conference on Computer Vision and Pattern Recognition}, 2020, pp. 4877--4886.

\bibitem{duzceker2021deepvideomvs}
A.~Duzceker, S.~Galliani, C.~Vogel, P.~Speciale, M.~Dusmanu, and M.~Pollefeys,
  ``{DeepVideoMVS}: Multi-view stereo on video with recurrent spatio-temporal
  fusion,'' in \emph{Proceedings of the IEEE/CVF Conference on Computer Vision
  and Pattern Recognition}, 2021, pp. 15\,324--15\,333.

\bibitem{long2021multi}
X.~Long, L.~Liu, W.~Li, C.~Theobalt, and W.~Wang, ``Multi-view depth estimation
  using epipolar spatio-temporal networks,'' in \emph{Proceedings of the
  IEEE/CVF Conference on Computer Vision and Pattern Recognition}, 2021, pp.
  8258--8267.

\bibitem{furukawa2015multi}
Y.~Furukawa and C.~Hern{\'a}ndez, ``Multi-view stereo: A tutorial,''
  \emph{Foundations and Trends{\textregistered} in Computer Graphics and
  Vision}, vol.~9, no. 1-2, pp. 1--148, 2015.

\bibitem{teed2020raft}
Z.~Teed and J.~Deng, ``{RAFT}: Recurrent all-pairs field transforms for optical
  flow,'' in \emph{European Conference on Computer Vision}.\hskip 1em plus
  0.5em minus 0.4em\relax Springer, 2020, pp. 402--419.

\bibitem{dai2017scannet}
A.~Dai, A.~X. Chang, M.~Savva, M.~Halber, T.~Funkhouser, and M.~Nie{\ss}ner,
  ``{ScanNet}: Richly-annotated {3D} reconstructions of indoor scenes,'' in
  \emph{Proceedings of the IEEE Conference on Computer Vision and Pattern
  Recognition}, 2017, pp. 5828--5839.

\bibitem{ummenhofer2017demon}
B.~Ummenhofer, H.~Zhou, J.~Uhrig, N.~Mayer, E.~Ilg, A.~Dosovitskiy, and
  T.~Brox, ``{DeMoN}: Depth and motion network for learning monocular stereo,''
  in \emph{Proceedings of the IEEE Conference on Computer Vision and Pattern
  Recognition}, 2017, pp. 5038--5047.

\bibitem{schops2017multi}
T.~Schops, J.~L. Schonberger, S.~Galliani, T.~Sattler, K.~Schindler,
  M.~Pollefeys, and A.~Geiger, ``A multi-view stereo benchmark with
  high-resolution images and multi-camera videos,'' in \emph{Proceedings of the
  IEEE Conference on Computer Vision and Pattern Recognition}, 2017, pp.
  3260--3269.

\bibitem{shotton2013scene}
J.~Shotton, B.~Glocker, C.~Zach, S.~Izadi, A.~Criminisi, and A.~Fitzgibbon,
  ``Scene coordinate regression forests for camera relocalization in {RGB-D}
  images,'' in \emph{Proceedings of the IEEE Conference on Computer Vision and
  Pattern Recognition}, 2013, pp. 2930--2937.

\bibitem{furukawa2009accurate}
Y.~Furukawa and J.~Ponce, ``Accurate, dense, and robust multiview stereopsis,''
  \emph{IEEE Transactions on Pattern Analysis and Machine Intelligence},
  vol.~32, no.~8, pp. 1362--1376, 2009.

\bibitem{lhuillier2005quasi}
M.~Lhuillier and L.~Quan, ``A quasi-dense approach to surface reconstruction
  from uncalibrated images,'' \emph{IEEE Transactions on Pattern Analysis and
  Machine Intelligence}, vol.~27, no.~3, pp. 418--433, 2005.

\bibitem{kar2017learning}
A.~Kar, C.~H{\"a}ne, and J.~Malik, ``Learning a multi-view stereo machine,'' in
  \emph{Proceedings of the 31st International Conference on Neural Information
  Processing Systems}, 2017, pp. 364--375.

\bibitem{ji2017surfacenet}
M.~Ji, J.~Gall, H.~Zheng, Y.~Liu, and L.~Fang, ``Surfacenet: An end-to-end {3D}
  neural network for multiview stereopsis,'' in \emph{Proceedings of the IEEE
  International Conference on Computer Vision}, 2017, pp. 2307--2315.

\bibitem{seitz1999photorealistic}
S.~M. Seitz and C.~R. Dyer, ``Photorealistic scene reconstruction by voxel
  coloring,'' \emph{International Journal of Computer Vision}, vol.~35, no.~2,
  pp. 151--173, 1999.

\bibitem{kutulakos2000theory}
K.~N. Kutulakos and S.~M. Seitz, ``A theory of shape by space carving,''
  \emph{International Journal of Computer Vision}, vol.~38, no.~3, pp.
  199--218, 2000.

\bibitem{stuhmer2010real}
J.~St{\"u}hmer, S.~Gumhold, and D.~Cremers, ``Real-time dense geometry from a
  handheld camera,'' in \emph{Joint Pattern Recognition Symposium}.\hskip 1em
  plus 0.5em minus 0.4em\relax Springer, 2010, pp. 11--20.

\bibitem{tola2012efficient}
E.~Tola, C.~Strecha, and P.~Fua, ``Efficient large-scale multi-view stereo for
  ultra high-resolution image sets,'' \emph{Machine Vision and Applications},
  vol.~23, no.~5, pp. 903--920, 2012.

\bibitem{pizzoli2014remode}
M.~Pizzoli, C.~Forster, and D.~Scaramuzza, ``Remode: Probabilistic, monocular
  dense reconstruction in real time,'' in \emph{2014 IEEE International
  Conference on Robotics and Automation (ICRA)}.\hskip 1em plus 0.5em minus
  0.4em\relax IEEE, 2014, pp. 2609--2616.

\bibitem{galliani2015massively}
S.~Galliani, K.~Lasinger, and K.~Schindler, ``Massively parallel multiview
  stereopsis by surface normal diffusion,'' in \emph{Proceedings of the IEEE
  International Conference on Computer Vision}, 2015, pp. 873--881.

\bibitem{yao2017relative}
Y.~Yao, S.~Li, S.~Zhu, H.~Deng, T.~Fang, and L.~Quan, ``Relative camera
  refinement for accurate dense reconstruction,'' in \emph{2017 International
  Conference on 3D Vision (3DV)}.\hskip 1em plus 0.5em minus 0.4em\relax IEEE,
  2017, pp. 185--194.

\bibitem{barnes2009patchmatch}
C.~Barnes, E.~Shechtman, A.~Finkelstein, and D.~B. Goldman, ``{PatchMatch}: A
  randomized correspondence algorithm for structural image editing,'' \emph{ACM
  Trans. Graph.}, vol.~28, no.~3, p.~24, 2009.

\bibitem{merrell2007real}
P.~Merrell, A.~Akbarzadeh, L.~Wang, P.~Mordohai, J.-M. Frahm, R.~Yang,
  D.~Nist{\'e}r, and M.~Pollefeys, ``Real-time visibility-based fusion of depth
  maps,'' in \emph{2007 IEEE 11th International Conference on Computer
  Vision}.\hskip 1em plus 0.5em minus 0.4em\relax IEEE, 2007, pp. 1--8.

\bibitem{newcombe2011kinectfusion}
R.~A. Newcombe, S.~Izadi, O.~Hilliges, D.~Molyneaux, D.~Kim, A.~J. Davison,
  P.~Kohi, J.~Shotton, S.~Hodges, and A.~Fitzgibbon, ``Kinectfusion: Real-time
  dense surface mapping and tracking,'' in \emph{2011 10th IEEE International
  Symposium on Mixed and Augmented Reality}.\hskip 1em plus 0.5em minus
  0.4em\relax IEEE, 2011, pp. 127--136.

\bibitem{zeng20173dmatch}
A.~Zeng, S.~Song, M.~Nie{\ss}ner, M.~Fisher, J.~Xiao, and T.~Funkhouser,
  ``{3DMatch}: Learning local geometric descriptors from {RGB-D}
  reconstructions,'' in \emph{Proceedings of the IEEE Conference on Computer
  Vision and Pattern Recognition}, 2017, pp. 1802--1811.

\bibitem{izadi2011kinectfusion}
S.~Izadi, D.~Kim, O.~Hilliges, D.~Molyneaux, R.~Newcombe, P.~Kohli, J.~Shotton,
  S.~Hodges, D.~Freeman, A.~Davison \emph{et~al.}, ``Kinectfusion: real-time
  3{D} reconstruction and interaction using a moving depth camera,'' in
  \emph{Proceedings of the 24th annual ACM symposium on User Interface Software
  and Technology}, 2011, pp. 559--568.

\bibitem{hou2019multi}
Y.~Hou, J.~Kannala, and A.~Solin, ``Multi-view stereo by temporal nonparametric
  fusion,'' in \emph{Proceedings of the IEEE/CVF International Conference on
  Computer Vision}, 2019, pp. 2651--2660.

\bibitem{luo2019p}
K.~Luo, T.~Guan, L.~Ju, H.~Huang, and Y.~Luo, ``P-{MVSNet}: Learning patch-wise
  matching confidence aggregation for multi-view stereo,'' in \emph{Proceedings
  of the IEEE/CVF International Conference on Computer Vision}, 2019, pp.
  10\,452--10\,461.

\bibitem{xue2019mvscrf}
Y.~Xue, J.~Chen, W.~Wan, Y.~Huang, C.~Yu, T.~Li, and J.~Bao, ``{MVSCRF}:
  Learning multi-view stereo with conditional random fields,'' in
  \emph{Proceedings of the IEEE/CVF International Conference on Computer
  Vision}, 2019, pp. 4312--4321.

\bibitem{yao2019recurrent}
Y.~Yao, Z.~Luo, S.~Li, T.~Shen, T.~Fang, and L.~Quan, ``Recurrent {MVSNet} for
  high-resolution multi-view stereo depth inference,'' in \emph{Proceedings of
  the IEEE/CVF Conference on Computer Vision and Pattern Recognition}, 2019,
  pp. 5525--5534.

\bibitem{sinha2020deltas}
A.~Sinha, Z.~Murez, J.~Bartolozzi, V.~Badrinarayanan, and A.~Rabinovich,
  ``{DELTAS}: Depth estimation by learning triangulation and densification of
  sparse points,'' in \emph{Computer Vision--ECCV 2020: 16th European
  Conference, Glasgow, UK, August 23--28, 2020, Proceedings, Part XXI
  16}.\hskip 1em plus 0.5em minus 0.4em\relax Springer, 2020, pp. 104--121.

\bibitem{gu2020cascade}
X.~Gu, Z.~Fan, S.~Zhu, Z.~Dai, F.~Tan, and P.~Tan, ``Cascade cost volume for
  high-resolution multi-view stereo and stereo matching,'' in \emph{Proceedings
  of the IEEE/CVF Conference on Computer Vision and Pattern Recognition}, 2020,
  pp. 2495--2504.

\bibitem{ke2020deep}
T.~Ke, T.~Do, K.~Vuong, K.~Sartipi, and S.~I. Roumeliotis, ``Deep multi-view
  depth estimation with predicted uncertainty,'' in \emph{2021 IEEE
  International Conference on Robotics and Automation (ICRA)}.\hskip 1em plus
  0.5em minus 0.4em\relax IEEE, 2021, pp. 9235--9241.

\bibitem{yan2020dense}
J.~Yan, Z.~Wei, H.~Yi, M.~Ding, R.~Zhang, Y.~Chen, G.~Wang, and Y.-W. Tai,
  ``Dense hybrid recurrent multi-view stereo net with dynamic consistency
  checking,'' in \emph{European Conference on Computer Vision}.\hskip 1em plus
  0.5em minus 0.4em\relax Springer, 2020, pp. 674--689.

\bibitem{schonberger2016structure}
J.~L. Schonberger and J.-M. Frahm, ``Structure-from-motion revisited,'' in
  \emph{Proceedings of the IEEE Conference on Computer Vision and Pattern
  Recognition}, 2016, pp. 4104--4113.

\bibitem{lindenberger2021pixel}
P.~Lindenberger, P.-E. Sarlin, V.~Larsson, and M.~Pollefeys, ``Pixel-perfect
  structure-from-motion with featuremetric refinement,'' in \emph{Proceedings
  of the IEEE/CVF International Conference on Computer Vision}, 2021, pp.
  5987--5997.

\bibitem{mur2017orb}
R.~Mur-Artal and J.~D. Tard{\'o}s, ``{ORB-SLAM2}: An open-source {SLAM} system
  for monocular, stereo, and {RGB-D} cameras,'' \emph{IEEE Transactions on
  Robotics}, vol.~33, no.~5, pp. 1255--1262, 2017.

\bibitem{mur2015orb}
R.~Mur-Artal, J.~M.~M. Montiel, and J.~D. Tardos, ``{ORB-SLAM}: A versatile and
  accurate monocular {SLAM} system,'' \emph{IEEE Transactions on Robotics},
  vol.~31, no.~5, pp. 1147--1163, 2015.

\bibitem{lowe2004distinctive}
D.~G. Lowe, ``Distinctive image features from scale-invariant keypoints,''
  \emph{International Journal of Computer Vision}, vol.~60, no.~2, pp. 91--110,
  2004.

\bibitem{rublee2011orb}
E.~Rublee, V.~Rabaud, K.~Konolige, and G.~Bradski, ``{ORB}: An efficient
  alternative to {SIFT} or {SURF},'' in \emph{2011 International Conference on
  Computer Vision}.\hskip 1em plus 0.5em minus 0.4em\relax Ieee, 2011, pp.
  2564--2571.

\bibitem{detone2018superpoint}
D.~DeTone, T.~Malisiewicz, and A.~Rabinovich, ``Superpoint: Self-supervised
  interest point detection and description,'' in \emph{Proceedings of the IEEE
  Conference on Computer Vision and Pattern Recognition Workshops}, 2018, pp.
  224--236.

\bibitem{liu2020extremely}
X.~Liu, Y.~Zheng, B.~Killeen, M.~Ishii, G.~D. Hager, R.~H. Taylor, and
  M.~Unberath, ``Extremely dense point correspondences using a learned feature
  descriptor,'' in \emph{Proceedings of the IEEE/CVF Conference on Computer
  Vision and Pattern Recognition}, 2020, pp. 4847--4856.

\bibitem{yang2008stereo}
Q.~Yang, L.~Wang, R.~Yang, H.~Stew{\'e}nius, and D.~Nist{\'e}r, ``Stereo
  matching with color-weighted correlation, hierarchical belief propagation,
  and occlusion handling,'' \emph{IEEE Transactions on Pattern Analysis and
  Machine Intelligence}, vol.~31, no.~3, pp. 492--504, 2008.

\bibitem{kanade1994stereo}
T.~Kanade and M.~Okutomi, ``A stereo matching algorithm with an adaptive
  window: Theory and experiment,'' \emph{IEEE Transactions on Pattern Analysis
  and Machine Intelligence}, vol.~16, no.~9, pp. 920--932, 1994.

\bibitem{dosovitskiy2015flownet}
A.~Dosovitskiy, P.~Fischer, E.~Ilg, P.~Hausser, C.~Hazirbas, V.~Golkov, P.~Van
  Der~Smagt, D.~Cremers, and T.~Brox, ``{FlowNet}: Learning optical flow with
  convolutional networks,'' in \emph{Proceedings of the IEEE International
  Conference on Computer Vision}, 2015, pp. 2758--2766.

\bibitem{long2020occlusion}
X.~Long, L.~Liu, C.~Theobalt, and W.~Wang, ``Occlusion-aware depth estimation
  with adaptive normal constraints,'' in \emph{European Conference on Computer
  Vision}.\hskip 1em plus 0.5em minus 0.4em\relax Springer, 2020, pp. 640--657.

\bibitem{xiao2013sun3d}
J.~Xiao, A.~Owens, and A.~Torralba, ``{SUN3D}: A database of big spaces
  reconstructed using {SfM} and object labels,'' in \emph{Proceedings of the
  IEEE International Conference on Computer Vision}, 2013, pp. 1625--1632.

\bibitem{aanaes2016large}
H.~Aan{\ae}s, R.~R. Jensen, G.~Vogiatzis, E.~Tola, and A.~B. Dahl,
  ``Large-scale data for multiple-view stereopsis,'' \emph{International
  Journal of Computer Vision}, vol. 120, no.~2, pp. 153--168, 2016.

\bibitem{Knapitsch2017}
A.~Knapitsch, J.~Park, Q.-Y. Zhou, and V.~Koltun, ``{Tanks and Temples}:
  Benchmarking large-scale scene reconstruction,'' \emph{ACM Transactions on
  Graphics}, vol.~36, no.~4, 2017.

\bibitem{sturm2012benchmark}
J.~Sturm, N.~Engelhard, F.~Endres, W.~Burgard, and D.~Cremers, ``A benchmark
  for the evaluation of {RGB-D} {SLAM} systems,'' in \emph{2012 IEEE/RSJ
  International Conference on Intelligent Robots and Systems}.\hskip 1em plus
  0.5em minus 0.4em\relax IEEE, 2012, pp. 573--580.

\bibitem{loshchilov2017decoupled}
I.~Loshchilov and F.~Hutter, ``Decoupled weight decay regularization,'' in
  \emph{International Conference on Learning Representations}, 2018.

\end{thebibliography}
%


 





\end{document}